\documentclass[11pt,a4paper]{article}
\usepackage[hyperref]{emnlp2018}
\usepackage{times}
\usepackage{latexsym}

\usepackage{url}

\usepackage{multirow}
\usepackage{graphicx}
\usepackage{subcaption}
\usepackage{acronym} 
\usepackage{todonotes}
\aclfinalcopy % Uncomment this line for the final submission

\newacro{pos}[PoS]{Part of Speech}
\newacro{bow}[BoW]{Bag of Words}
\newacro{svm}[SVM]{support vector machines}

\title{DepecheMood++: a Bilingual Emotion Lexicon Built Through Simple Yet Powerful Techniques}

\author{Oscar Araque$^1$, Lorenzo Gatti$^2$, Jacopo Staiano$^3$, Marco Guerini$^{4,5}$ \\
  $^1$Grupo de Sistemas Inteligentes, Departamento de Ingenier\'ia Telem\'atica\\
Universidad Polit\'ecnica de Madrid, E.T.S.I. de Telecomunicaci\'on, Madrid -- Spain \\
  $^2$ Human Media Interaction, University of Twente,
Enschede -- The Netherlands\\
  $^3$ reciTAL, 34 boulevard de Bonne Nouvelle,
75010 Paris -- France\\
  $^4$ Fondazione Bruno Kessler, Via Sommarive 18, Povo, Trento -- Italy\\
  $^5$ AdeptMind Scholar -- Canada\\
  {\tt o.araque@upm.es}, {\tt l.gatti@utwente.nl}, {\tt jacopo@recital.ai}, {\tt guerini@fbk.eu}\\
}  

\date{}

\begin{document}
\maketitle
\begin{abstract}
Several lexica for sentiment analysis have been developed and made available in the NLP community. While most of these come with word polarity annotations (e.g. positive/negative), attempts at building lexica for finer-grained emotion analysis (e.g. happiness, sadness) have recently attracted significant attention. Such lexica are often exploited as a building block in the process of developing learning models for which emotion recognition is needed, and/or used as baselines to which compare the performance of the models.
In this work, we contribute two new resources to the community: a) an extension of an existing and widely used emotion lexicon for English; and b) a novel version of the lexicon targeting Italian. Furthermore, we show how simple techniques can be used, both in supervised and unsupervised experimental settings, to boost  performances on datasets and tasks of varying degree of domain-specificity.
\end{abstract}

\section{Introduction}

Obtaining high-quality and high-coverage lexica is an active subject of research~\cite{mohammad2010emotions}.
Traditionally, lexicon acquisition can be done in two distinct ways: either manual creation (e.g. crowdsourcing annotation) or automatic 
derivation from already annotated corpora. While the former approach provides more precise lexica, the latter usually grants a higher coverage. Regardless of the approach chosen, when used as baselines or as additional features for learning models, lexica are often ``taken for granted", meaning that the performances against which a proposed model is evaluated are rather weak, a fact that could be arguably seen to slow down progress in the field. Thus, in this paper we first investigate whether simple and computationally cheap techniques (e.g. document filtering, text pre-processing, frequency cut-off) can be used to improve both \textit{precision} and \textit{coverage} of a state-of-the-art lexicon  that has been automatically inferred from a dataset of emotionally tagged news. 
Then, we try to answer the following research questions:
\begin{itemize}
    \item Can straightforward  machine learning techniques that only rely on lexicon scores provide even more challenging baselines for complex emotion analysis models, under the constraints of keeping the required pre-processing at a minimum?
    \item Are such techniques portable across languages?
    \item Can the coverage of a given lexicon be significantly increased using a straight-forward and effective methodology?
    %\item \textcolor{violet}{When used as additional features in complex learning models, can the enhanced lexica improve performance of such models? }
\end{itemize}

To do so, we build upon the methodology proposed in~\cite{staiano2014depechemood,guerini2015deep}, the publicly available \texttt{DepecheMood} lexicon described therein, and the corresponding details of the source dataset we were provided with. 

We evaluate and release to the community an extension of the original lexicon built on a larger dataset, as well as a novel emotion lexicon targeting the Italian language and built with the same methodology. %\footnote{Lexica and code for the experiments presented in this paper can be found in the supplementary material. We commit to publicly release those upon publication of this work.}. 
We perform experiments on six datasets/tasks exhibiting a wide diversity in terms of \textit{domain} (namely: news, blog posts, mental health forum posts, twitter), \textit{languages} (English, and Italian), \textit{setting} (both supervised and unsupervised), and \textit{task} (regression and classification). 

The results obtained show that: 
\begin{enumerate}
    \item training straightforward classifiers/regressors from a high-coverage/high-precision lexicon, derived from general news data, allows to obtain good performances also on domain-specific tasks, and provides more challenging baselines for complex task-specific models;
    \item depending on the characteristics of the target language, specific pre-processing steps (e.g. lemmatization in case of morphologically-rich languages) can be beneficial;
    \item coverage of the original lexicon can be extended using embeddings, and such technique can provide performance improvements.
\end{enumerate}

\section{Related Work}
Here we provide a short review of efforts towards building sentiment and emotion lexica; the interested reader, can find a more thorough  overview in~\cite{pang2008opinion,liu2012survey,wilson:AAAI-04,Paltoglou2010}.

\begin{table*} [htb!] 	
	\setlength{\tabcolsep}{4pt} %shrink column size
	\begin{center} 	 	
		{\footnotesize 		
			\begin{tabular}{llcccc} 		
				\hline 

	\textbf{Domain} & \textbf{Name} & \textbf{No. entries} & \textbf{Annotation type} & \textbf{Annotation process} & \textbf{Reference}  \\
    \hline
    \multirow{8}{*}{Sentiment} &  General Inquirer  & 4k & Categorical & Manual &  \cite{stone1966general} \\
    & ANEW  & 1k & Numerical & Manual &  \cite{bradley1999affective} \\
    & -  & 6k & Categorical & Automatic &  \cite{hu2004mining} \\
    & SentiWordNet  & 117k & Numerical & Automatic &  \cite{baccianella2010sentiwordnet} \\
    & AFINN  & 2k & Numerical & Manual &  \cite{nielsen2011new} \\
    & SO-CAL  & 4k & Numerical & Manual &  \cite{taboada2011lexicon} \\
    & -  & 14k & Numerical & Manual &  \cite{warriner2013norms} \\
    & SentiWords  & 144k & Numerical & Automatic &  \cite{gatti2016sentiwords} \\
    & NRC VAD & 20k & Numerical & Manual & \cite{vad-acl2018} \\
    \hline
    \multirow{7}{*}{Emotion} & Fuzzy Affect & 4k & Numerical & Manual &  \cite{subasic2001affect} \\
    & WordNet-Affect & 1k & Categorical & Manual &  \cite{strappaLREC04} \\
    & Affect database & 2k & Numerical & Manual &  \cite{neviarouskaya2010emoheart} \\
    & AffectNet & 10k & Categorical & Automatic &  \cite{cambria2011sentic} \\
    & EmoLex & 14k & Categorical & Manual &  \cite{Mohammad13} \\
    & NRC Hashtag & 16k & Numerical & Automatic & \cite{COIN12024} \\
    & NRC Affect  & 6k & Numerical & Manual & \cite{LREC18-AIL} \\
    \hline
	\end{tabular} 		
	} 		 	
	\end{center}	 	
	\setlength{\belowcaptionskip}{-0.1cm} 	
	\caption{Overview of related Sentiment and Emotion lexica.} 
	\label{tab:lexica-reference}
\end{table*}

\subsection{Sentiment Lexica}

A number of sentiment lexica have been developed during the years, with considerable differences in the number of annotated words (from a thousand to hundreds of thousands), the values they associate to a single word (from binary to multi-class, to finer-grained scales), and in the way these ratings are collected (manually or automatically). Here, we only report some notable and accessible examples. 

\texttt{General Inquirer} \cite{stone1966general} is one of the earliest resources of such kind, and provides binary ratings for about 4k sentiment words, as well as a number of syntactic, semantic, and pragmatic categories.
More than three times larger, the resource by \citet{warriner2013norms} provides fine-grained ratings for 14k frequent-usage words, obtained by averaging the crowdsourced answers of multiple annotators.
This dataset is an extension of the Affective Norms for English Words (\texttt{ANEW}), which reports similar scores for a set of 1k words~\cite{bradley1999affective}.
It is worth noting that \texttt{ANEW} valence scores have been manually assigned by several annotators, leading to an increase in precision.

Following the \texttt{ANEW} methodology, a microblogging-oriented resource has been introduced by \citet{nielsen2011new}, called \texttt{AFINN}.
Its latest version comprises 2477 words and phrases that have been manually annotated.
As shown by the original author, the precision of the \texttt{AFINN} resource, in comparison to other lexica, can be higher when applied to analysis of microblogging platforms.
Similarly, \texttt{SO-CAL} \cite{taboada2011lexicon} entries have been generated by a small group of human annotators.
Such annotation has been made following a multi-class approach, obtaining a finer resolution in the valence scores, which range from -5 (very negative) to 5 (very positive); further, these valence scores have been subsequently validated using crowd-sourcing, with the final size of the resource compounding to over 4k words.

Another relevant resource is \texttt{SentiWordNet} \cite{baccianella2010sentiwordnet}, which has been generated from a few seed terms to annotate each word sense of WordNet \cite{wordnet} with both a positive and negative score, as well as an objectivity score, in the $[0,1]$ range.
Building on top of it, the \texttt{SentiWords} resource \cite{gatti2016sentiwords} has been generated by used machine learning to improve the precision of these scores, annotating all the 144k lemmas of WordNet: taking into account the valence expressed in manually annotated lexica, the method proposed is based on predicting the valence score of previously unseen words.

Several works in the literature makes use of the lexicon presented in \cite{hu2004mining}: this dictionary consists of more than 6k words, including frequent sentiment words, slang words, misspelled terms, and common variants.
The annotations are automated using adjective words as seed, and expanding the valence value using synonym and antonym relations between words, as expressed in WordNet.
A recent work \cite{vad-acl2018}, called \texttt{NRC Valence, Arousal, Dominance Lexicon} contains 20k terms annotated with valence, arousal and dominance: the proposed generation process relies on a method known as Best-Worst Scaling \cite{bws-naacl2016}, which aims at avoiding some issues common to human annotators.

\subsection{Emotion Lexica}
While many sentiment lexica have been produced, fewer linguistic resources for emotion research are described in the literature. Among these, a known resource is 
\texttt{WordNet-Affect}~\cite{strappaLREC04}, a manually-built extension of WordNet, in which about 1k lemmas are assigned with a label taken from a hierarchy of 311 affective labels, including Ekman's six emotions~\cite{Ekman1971}. \texttt{AffectNet}~\cite{cambria2011sentic} is a semantic network containing about 10k items, created by blending entries from ConceptNet \cite{havasi2007conceptnet} and the emotional labels of \texttt{WordNet-Affect}. 

Similarly, the \texttt{Affect} database \cite{neviarouskaya2010emoheart} contains 2.5k lemmas taken from \texttt{WordNet-Affect}, and has been manually enriched by adding the strength of association with Izard's basic emotions \cite{izard1977human}.
\texttt{EmoLex} \cite{Mohammad13} is a crowdsourced lexicon containing 14k lemmas, each annotated with binary associations to  Plutchik's eight emotions~\cite{plutchik1980general}.
%\texttt{DepecheMood}~\cite{staiano2014depechemood},  provided ratings across 8 emotions for 37k items, harvesting them from human ratings of articles on social news networks.

Further, a fuzzy approach is considered by \citet{subasic2001affect}, who provide 4k entries manually annotated in a range of 80 emotion labels.
A recent resource is \texttt{NRC Affect Intensity Lexicon} \cite{LREC18-AIL}, which includes 6k entries manually annotated with a set of four basic emotions: joy, fear, anger, and sadness.
For this lexicon, similarly to before, the Best-Worst Scaling method was used.
In a similar line of work has the \texttt{NRC Hashtag Emotion Lexicon} \cite{COIN12024} as output.
With a coverage of over 16k unigrams, this resource has been automatically inferred from microblogging messages distantly annotated by emotional hashtags.
As such, this lexicon is particularly useful when applied to the Twitter domain.

\section{DepecheMood++}
\label{sec:depechemood++}

In this section we provide details on the techniques and datasets we used to create \texttt{DepecheMood++} (\texttt{DM++} for short), an extension of the  \texttt{DepecheMood} lexicon (which from now on we will refer to as \texttt{DepecheMood$_{2014}$}, or \texttt{DM$_{2014}$}). The original lexicon~\cite{staiano2014depechemood}, built in a completely automated and domain-agnostic fashion, has been extensively used by the research community and has demonstrated high performance even in domain-specific tasks, often outperformed only by domain-specific lexica/systems; see for instance ~\cite{bobicev2015goes}.

%\textcolor{red}{REMOVE?: To generate the  \texttt{DepecheMood++} (\texttt{DM++}) lexica, we have replicated the approach presented by~\cite{staiano2014depechemood}, REMOVE ONLY THE FOLLOWING? 
%where word-by-emotion and word-by-document matrices are built, and through matrix to matrix product the word-by-emotion matrix is computed.}
%\textcolor{red}{Also, based on the findings of the original work, and to reduce computational cost, we evaluate the \textit{normalized} version of the resource. JS: THIS COULD BE CUT}

% \textcolor{violet}{We choose this lexicon since it proved to be the SOTA lexicon for emotion recognition \cite{staiano2014depechemood} and also proved to be the best generic lexicon in domain specific tasks \textbf{[CIT]}.} 
The new version we release in this work is made available for both English and Italian. While the English version of \texttt{DM++} is an improved version of \texttt{DM$_{2014}$} built using a larger dataset, the Italian one is completely new and, to the best of our knowledge, is the first publicly-available large-scale emotion lexicon for this language. 

%and on the resources used to evaluate the new lexica performances.

\subsection{Data Used}

%To build the \texttt{DM++} lexica, we followed the methodology presented in~\cite{staiano2014depechemood}, where the the authors harvested all the news articles from \texttt{rappler.com}, as of June 3rd 2013. 

To build \texttt{DepecheMood$_{2014}$}, the original authors exploited a dataset consisting of 13.5M words from 25.3K documents, with an average of 530 words per document.

As previously mentioned, in this paper we use an expanded source dataset in order to \emph{i)} re-build the English lexicon on a larger corpus, and \emph{ii)} to build a novel lexicon targeting the Italian language. To this end, we used an extended corpus which has been harvested for a subsequent study on emotions and virality~\cite{guerini2015deep} -- besides the English articles from \texttt{rappler.com} on a longer time span, such corpus includes crowd-annotated news articles in Italian from \texttt{corriere.it}.

% \oa{I'm removing figures of the votes distribution.}
In brief, \texttt{rappler.com} is a ``social-news'' website that embeds a small interface, called \emph{Mood Meter}, in every article it publishes, allowing its readers to express with a click their emotional reaction to the story they are reading.
%The percentages of votes obtained by each emotion are also visualized by the interface, which is depicted in Figure~\ref{fig:moodmeter}.\\
Similarly, \texttt{corriere.it}, the online version of a very popular Italian newspaper called \emph{Corriere della Sera}, adopts a similar approach, based on \emph{emoticons}, to sense the emotional reactions of its readers. We note that the latter has discontinued its ``emotional'' widgets, removing them also from the archived articles, a fact that contributes to the relevance of our effort to release the Italian \texttt{DM++} lexicon.% in this work. 
In Table~\ref{tab:corpora-description} we report a quantitative description of the data collected from \texttt{Rappler} and \texttt{Corriere}.
For more details, we refer the reader to the original work of~\citet{guerini2015deep}.

\begin{table} [!htb] 	
	\begin{center} 	 	
		{\footnotesize 		
			\begin{tabular}{l|cc}
				\hline
        & \textbf{Rappler} & \textbf{Corriere} \\
        \hline 
        Documents  & 53,226 & 12,437 \\
        Tot. words & 18.17 M & 4.04 M \\
        %Emotional Votes & 1,145,543* & 210,113\\
        Words per Document & 341 & 325 \\
        No. of annotations & 1,145,543 & 320,697 \\
        \hline
		\end{tabular} 		
		} 		 	
	\end{center}	 	
	\setlength{\belowcaptionskip}{-0.1cm} 	
	\caption{Corpus statistics}
	\label{tab:corpora-description} 
\end{table}

% \oa{Can we get the votes info? If not, i would just delete that row. JS: don't think so, we can only estimate a lower bound as reported in the WWW paper ("emotional Votes"), so lets remove it - also, we should specify better what "WordsxDocument" is (avg? median?).}
%\textcolor{violet}{REMOVE IF SPACE IS NEEDED: As noted in~\cite{guerini2015deep}, the actual absolute number of votes collected by the widgets cannot be known, but a very conservative estimate for it can be computed: using the lower common denominator over the percentages of affective votes obtained by a Rappler article, the minimum number of votes needed to obtain them can be derived: such figure compounds to 1,145,543 votes over the \texttt{rappler.com} dataset, and 210,113 votes on the \texttt{corriere.it} dataset.}

% \begin{figure}[ht!]
% \centering
% \includegraphics[width=50mm]{RapplerWorkingTheCrowd_mod.png}
% \caption{A simple caption}
% \label{mood-meter}
% \end{figure}

While previous research efforts have exploited documents with emotional annotations on various affect-related tasks~\cite{mishne2005experiments,strapparava2008learning,bellegarda2010emotion,tang2014building}, the data used in these works share the limitation of only providing discrete labels,
% (\emph{i.e.} the document being a tweet, a blogpost or similar, and the emotion label a meta-data or an emoticon within the text)
rather than a continuous score for each emotional dimension. Moreover, these annotations were performed by the document author rather than the readers.

Conversely, in this work we leverage the fact that \texttt{rappler.com} and  \texttt{corriere.it} readers can select as many emotions as desired, so that the resulting annotations represent a distribution of emotional scores for each article.

\subsection{Emotion Lexica Creation}

Consistently with \citet{staiano2014depechemood} the lexica creation methodology consists of the following steps:
\begin{enumerate}
    \item First, we produced a document-by-emotion matrix ($M_{DE}$) per language, containing the voting percentages for each document 
in the eight affective dimensions available in \texttt{rappler.com} for English and the six available in \texttt{corriere.it} for Italian.
    \item Then, we computed the word-by-document matrices using normalized frequencies ($M_{WD}$).
    \item After that, we applied matrix multiplication between the document-by-emotion and word-by-document matrices ($M_{DE} \cdot M_{WD}$) to obtain a (raw) word-by-emotion matrix $M_{WE}$. This method allows us to `merge' words with emotions by summing the products of the weight of a word with the weight of the emotions in each document.
\end{enumerate}

Finally, we transformed $M_{WE}$ 
% in a \emph{probability} matrix 
by first applying 
%mean normalization on the columns to each sentiment score 
normalization column-wise
(so to eliminate the over representation for happiness as discussed in previous section) 
and then scaling the data row-wise so to sum up to one.

\begin{table*} [!htb] 	
	\begin{center} 	 	
		{\footnotesize 		
			\begin{tabular}{l|rrrrrrrrr} 	

\hline
 Word & \textsc{Afraid} & \textsc{Amused} & \textsc{Angry} & \textsc{Annoyed} & \textsc{Don't Care} & \textsc{HAppy} & \textsc{Inspired} & \textsc{Sad} \\
 \hline
awe & 0.04 & 0.18 & 0.04 & 0.12 & 0.14 & 0.12 & \textbf{0.32} & 0.04 \\
criminal & 0.08 & 0.11 & \textbf{0.27} & 0.16 & 0.11 & 0.09 & 0.07 & 0.10 \\
dead & 0.17 & 0.07 & 0.18 & 0.08 & 0.07 & 0.05 & 0.08 & \textbf{0.29} \\
funny & 0.04 & \textbf{0.31} & 0.04 & 0.13 & 0.17 & 0.09 & 0.16 & 0.06 \\
warning & \textbf{0.30} & 0.07 & 0.13 & 0.12 & 0.08 & 0.07 & 0.06 & 0.16 \\
rapist & 0.09 & 0.08 & \textbf{0.37} & 0.08 & 0.18 & 0.08 & 0.06 & 0.07 \\
virtuosity & 0.00 & 0.24 & 0.00 & 0.01 & 0.00 & \textbf{0.41} & 0.33 & 0.01 \\
\hline
\end{tabular} 		
		} 	
				 	
	\end{center}	 	
	\setlength{\belowcaptionskip}{-0.1cm} 	
	\caption{An excerpt of $M_{WE}$ for English. Dominant Emotion in a word is highlighted for readability purposes.} 	
	\label{tab:word-emotion_EN} 
\end{table*} 

\begin{table*} [!htb] 	
	\begin{center} 	 	
		{\footnotesize 		
			\begin{tabular}{l|rrrrrrrrr} 		
\hline
 & \textsc{Annoyed} & \textsc{Afraid} & \textsc{Sad} & \textsc{Amused} & \textsc{Happy} \\
\hline 
stupore & 0.18 & 0.16 & 0.16 & \textbf{0.28} & 0.22 \\
criminale & 0.21 & \textbf{0.28} & 0.16 & 0.20 & 0.15 \\
morto & 0.19 & 0.19 & \textbf{0.39} & 0.12 & 0.11 \\
divertente & 0.10 & 0.13 & 0.15 & \textbf{0.32} & 0.30 \\
allarme & 0.16 & 0.25 & \textbf{0.34} & 0.14 & 0.11 \\
stupratore & \textbf{0.40} & 0.11 & 0.16 & 0.18 & 0.15 \\
virtuoso & 0.14 & 0.18 & 0.14 & 0.25 & \textbf{0.29} \\

\hline
\end{tabular} 		
		} 	
				 	
	\end{center}	 	
	\setlength{\belowcaptionskip}{-0.1cm} 	
	\caption{An excerpt of $M_{WE}$ for Italian. Dominant Emotion in a word is highlighted for readability purposes.} 	
	\label{tab:word-emotion_IT} 
\end{table*}

An excerpt of the final Matrices $M_{WE}$ both for English and Italian are presented in Tables~\ref{tab:word-emotion_EN} and~\ref{tab:word-emotion_IT}: they can be interpreted as a list of words with scores that represent how much weight a given word has in each affective dimension.
These matrices, that we call \texttt{DepecheMood++}\footnote{%Reminiscent of the \emph{Depeche Mode} electronic band, 
In French, `depeche' means dispatch/news.}, represents our emotion lexica for English and Italian, and are freely available for research purposes at \url{https://git.io/fxGAP}. 

\subsection{Validation and Optimization}
\label{subsec:unsupervised}

In this section we describe several configurations that were used to generate \texttt{DepecheMood++} lexicon. To fairly assess the performance of each configuration, we employ randomly selected validation sets compounding to 25\% of the articles in our data for both \texttt{rappler.com} and \texttt{corriere.it}. For all the following evaluation experiments, such left-out sets are used. 

Also, in order to facilitate comparisons with previous works, we used the simple approach adopted both by~\citet{staiano2014depechemood} and \citet{strapparava2008learning}: on a given headline, a single value
for each affective dimension is computed by simply averaging the \texttt{DepecheMood++} affective scores of all the words contained in the headline;
Pearson correlation is then measured by comparing this averaged value to the annotation for the headline.

\paragraph{Word Representations.} Throughout the following experiments, we consider three word representations, corresponding to three different pre-processing levels: (i) tokenization, (ii) lemmatization, and (iii) lemmatization combined with \ac{pos} tagging (the only representation used in \texttt{DM$_{2014}$}).
We denote these variations as \texttt{token}, \texttt{lemma} and \texttt{lemma\#PoS} respectively.

\paragraph{Untagged Document Filtering.}  First, we noted that a significant percentage of the training document set has no emotional annotations -- such figure compounds to 8\% and 16\% for \texttt{rappler.com} and \texttt{corriere.it}, respectively. Therefore, we compare two versions of the proposed lexica built using two different sets of documents: (i) all documents, as done in the original \texttt{DM$_{2014}$}; and (ii), using only documents with a non-zero emotion annotation vector.

The results of this experiment are shown in Table~\ref{tab:emotional_documents}.

\begin{table} [htb!] 	
	\begin{center} 	 	
		{\footnotesize 		
			\begin{tabular}{l|ccc} 		
				\hline 

	\textbf{Dataset} & \textbf{token}  &  \textbf{lemma} & \textbf{lemma\#PoS}  \\
    \hline
    Rappler (all) & 0.31 & 0.31 & 0.30 \\
    Rappler (filtered) & \textbf{0.33} & \textbf{0.32} & \textbf{0.32} \\ 
    \hline
    Corriere (all) & 0.22 & 0.25 & 0.24 \\
    Corriere (filtered) & \textbf{0.27} & \textbf{0.30} & \textbf{0.30} \\ 
    \hline
	\end{tabular} 		
	} 		 	
	\end{center}	 	
	\setlength{\belowcaptionskip}{-0.1cm} 	
	\caption{Averaged Pearson's correlation over all emotions on left-out sets for Rappler and Corriere, using all documents in the datasets vs filtering out those without emotional annotations.} 
	\label{tab:emotional_documents}
\end{table}  

It is evident that training with only documents with emotion annotations leads to an improvement of results, which could indicate that untagged documents add noise to the lexicon generation process.
Consequently, we use this improved variant in next experiments.

\paragraph{Frequency Cutoff.} We also explore different word frequency cutoff values to find a threshold that would remove noisy items without eliminating informative ones (in \texttt{DM$_{2014}$} no cutoff was used, hence hapax were also included in the vocabulary).
The performance of \texttt{DepecheMood++} under different cutoff values is reported in Table~\ref{tab:cutoffs-performance}: the best performance was obtained using a cutoff value of 10 for both Rappler and Corriere. Therefore, we use this value on the following experiments. In Table~\ref{tab:cutoffs-coverage} we also report the vocabulary size as a function of cutoff values.

\begin{table} [htb!] 	
	\begin{center} 	 	
	\setlength{\tabcolsep}{4pt} %shrink column size
		{\footnotesize 		
			\begin{tabular}{c|ccc|ccc} 		
				\hline 

	\textbf{Cutoff} & \multicolumn{3}{c|}{\textbf{Rappler}} & \multicolumn{3}{c}{\textbf{Corriere}} \\
	\hline\hline
	& \textbf{token}  &  \textbf{lemma} & \textbf{l\#p} & \textbf{token}  &  \textbf{lemma} & \textbf{l\#p} \\
	\hline
    1   & 0.33 & 0.32 & 0.31 & 0.26 & 0.30 & 0.29 \\
    10  & 0.33 & 0.33 & 0.33 & 0.27 & 0.31 & 0.30 \\
    20  & 0.33 & 0.33 & 0.32 & 0.25 & 0.30 & 0.29 \\
    50  & 0.33 & 0.32 & 0.31 & 0.21 & 0.28 & 0.27 \\
    100 & 0.31 & 0.31 & 0.30 & 0.16 & 0.24 & 0.24 \\
    \hline
	\end{tabular} 		
	} 		 	
	\end{center}	 	
	\setlength{\belowcaptionskip}{-0.1cm} 	
	\caption{Frequency cutoff impact on Pearson's correlation for left-out sets for Rappler and Corriere.} 
	\label{tab:cutoffs-performance}
\end{table}

\begin{table} [htb!] 	
	\begin{center} 	 	
	\setlength{\tabcolsep}{4pt} %shrink column size
		{\footnotesize 		
			\begin{tabular}{c|ccc|ccc} 		
				\hline 

	\textbf{Cutoff} & \multicolumn{3}{c|}{\textbf{Rappler}} & \multicolumn{3}{c}{\textbf{Corriere}} \\
	\hline\hline
	& \textbf{token}  &  \textbf{lemma} & \textbf{l\#p} & \textbf{token}  &  \textbf{lemma} & \textbf{l\#p} \\
	\hline
    1   & 165k & 154k & 249k & 116k & 72k & 81k \\
    10  & 37k & 30k & 44k & 20k & 13k & 13k \\
    20  & 26k & 20k & 29k & 12k & 8k & 8k \\
    50  & 16k & 12k & 16k & 6k & 4k & 4k \\
    100 & 10k & 8k & 10k & 3k & 3k & 3k \\
    \hline
	\end{tabular} 		
	} 		 	
	\end{center}	 	
	\setlength{\belowcaptionskip}{-0.1cm} 	
	\caption{Number of words in generated lexica using different cutoff values.} 
	\label{tab:cutoffs-coverage}
\end{table}

% \begin{table} [htb!] 	
% 	\begin{center} 	 	
% 		{\footnotesize 		
% 			\begin{tabular}{l|ccc} 		
% 				\hline 

% 	\textbf{Dataset} & \textbf{token}  &  \textbf{lemma} & \textbf{lemma\#PoS}  \\
%     \hline
%     Rappler              & \multirow{2}{*}{0.35} & \multirow{2}{*}{0.35} & \multirow{2}{*}{0.34} \\
%     \hspace{1em} all \\
%     Rappler              & \multirow{2}{*}{0.38} & \multirow{2}{*}{0.37} & \multirow{2}{*}{0.36} \\
%     \hspace{1em} only emotion \\
%     \hline
%     Corriere              & \multirow{2}{*}{0.23} & \multirow{2}{*}{0.26} & \multirow{2}{*}{0.25} \\
%     \hspace{1em} all \\
%     Corriere             & \multirow{2}{*}{0.27} & \multirow{2}{*}{0.30} & \multirow{2}{*}{0.29} \\
%     \hspace{1em} only emotion \\
%     \hline
% 	\end{tabular} 		
% 	} 		 	
% 	\end{center}	 	
% 	\setlength{\belowcaptionskip}{-0.1cm} 	
% 	\caption{Comparison of averaged Pearson's correlation over all emotions on all datasets when filtering non-emotional documents.} 
% 	\label{tab:emotional_documents}
% \end{table}  

\begin{figure*}[!htb]
  \begin{subfigure}[b]{0.5\textwidth}
    \includegraphics[width=\textwidth]{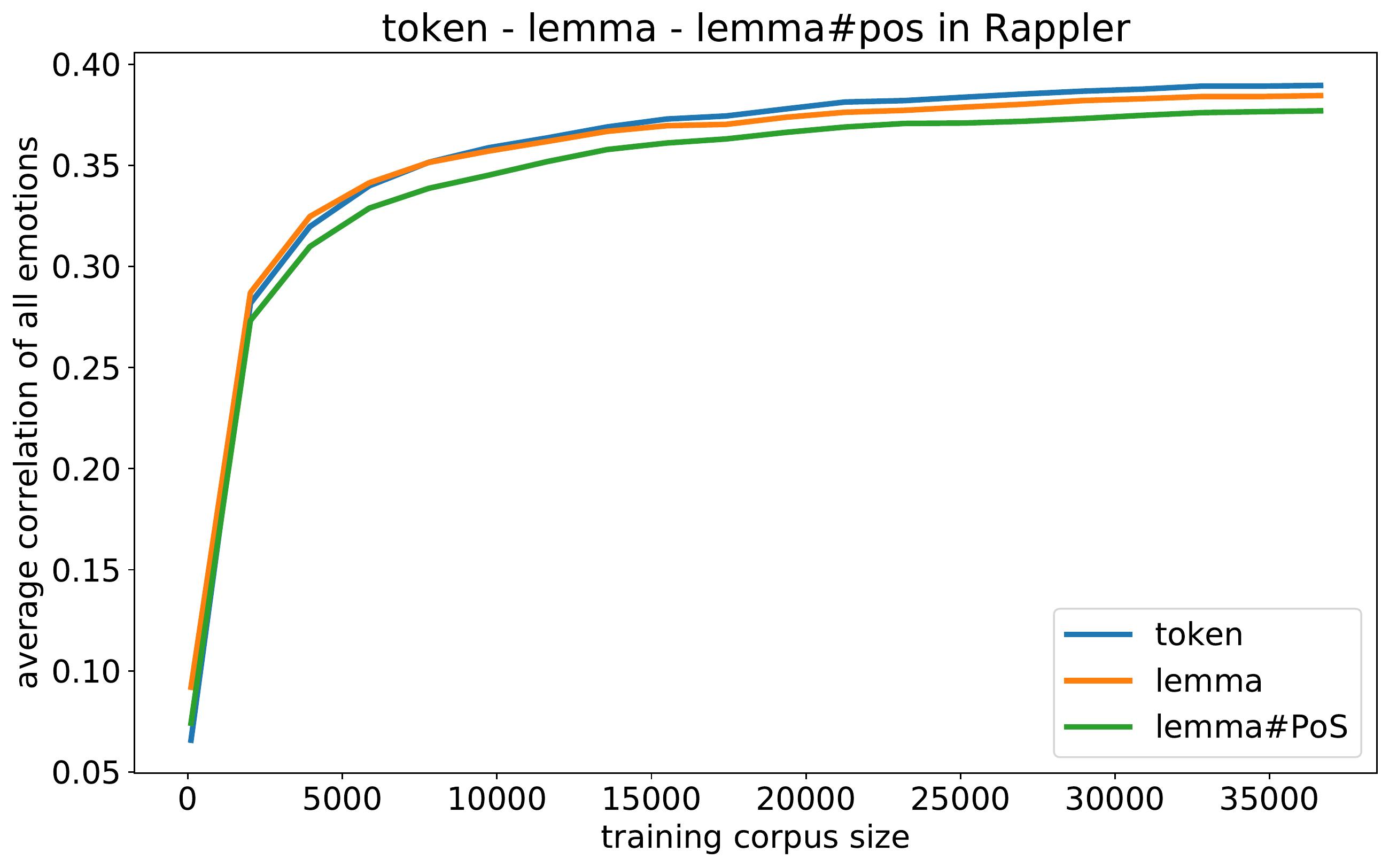}
    \caption{Rappler}
  \end{subfigure}
  \begin{subfigure}[b]{0.5\textwidth}
    \includegraphics[width=\textwidth]{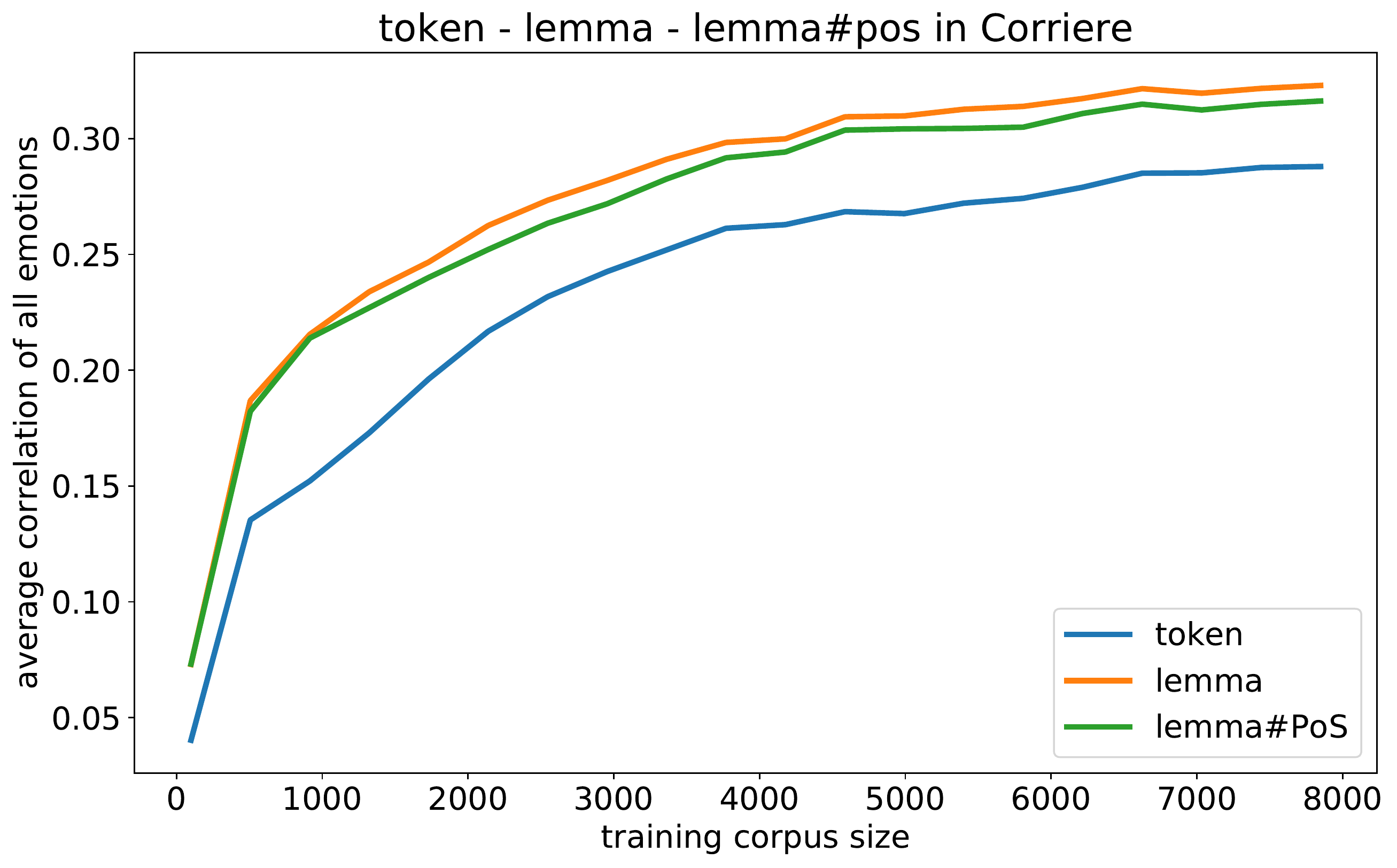}
    \caption{Corriere}
  \end{subfigure}
  
    \caption{Token - lemma - lemma\#PoS comparison in both \texttt{Rappler} and \texttt{Corriere}.}
    \label{fig:tokenlemma}
\end{figure*}

\paragraph{Learning Curves.} Next, we aim to understand if there is a limit, in terms of training dataset size, after which the performance saturates (indicating that further expansions of the corpus would not result beneficial). To this end, we vary the amount of documents used to build the lexica using the three different text pre-processing strategies (tokens, lemmatization and lemmatization with \ac{pos}) and evaluate their performance.

Figure~\ref{fig:tokenlemma} shows the correlation values on the left-out sets, yielded by lexica built upon training subsets of increasing size -- documents included at each subsequent step have been randomly selected from the original training sets.

The results show that, for the \texttt{Rappler} dataset, tokenization and lemmatization approaches consistently achieve the higher performance across various dataset sizes; conversely, on the \texttt{Corriere} dataset the lemmatization-based strategies yield the best performance with a significant improvement over tokenization.
A possible explanation for such performance drop can be hypothesized in the fact that the Italian language (\texttt{Corriere} data is in Italian) is morphologycally richer than English (as in the adjective \textit{good}, that can be written as \textit{buono} or \textit{buona}, \textit{buoni}, \textit{buone}); thus, lemmatization can reduce data sparseness that harms the final lexicon quality.

\begin{table} [htb!] 	
	\begin{center} 	 	
		{\footnotesize 		
			\begin{tabular}{l|c|ccc} 		
				\hline 
	\textbf{Emotion} & \textbf{DM$_{2014}$} &  \textbf{token}  &  \textbf{lemma} & \textbf{l\#p}  \\
    \hline
    \textsc{Afraid}      & 0.30 & 0.38 & \textbf{0.39} & 0.38 \\
    \textsc{Amused}      & 0.16 & \textbf{0.33} & 0.32 & 0.31 \\
    \textsc{Angry}       & 0.40 & 0.39 & \textbf{0.40} & \textbf{0.40} \\
    \textsc{Annoyed}     & 0.15 & \textbf{0.21} & \textbf{0.21} & \textbf{0.21} \\
    \textsc{Don't Care}  & 0.16 & 0.21 & \textbf{0.22} & 0.21 \\
    \textsc{Happy}       & 0.30 & \textbf{0.35} & \textbf{0.35} & \textbf{0.35} \\
    \textsc{Inspired}    & 0.31 & \textbf{0.36} & \textbf{0.36} & 0.35 \\
    \textsc{Sad}         & 0.36 & 0.39 & \textbf{0.41} & \textbf{0.41} \\
    \hline
    \textsc{Average}    & 0.27 & \textbf{0.33} & \textbf{0.33} & \textbf{0.33} \\
    \hline 	
			\end{tabular} 		
		} 		 	
	\end{center}	 	
	\setlength{\belowcaptionskip}{-0.1cm} 	
	\caption{Pearson's correlation on \texttt{Rappler} left-out set.}
	\label{tab:Rapplerdev}  
\end{table}  

\begin{table} [htb!] 	
	\begin{center} 	 	
		{\footnotesize 		
			\begin{tabular}{l|ccc} 		
				\hline 

	\textbf{Emotion} & \textbf{token}  &  \textbf{lemma} & \textbf{lemma\#PoS}  \\
    \hline
    \textsc{Annoyed} & 0.40 & \textbf{0.44} & \textbf{0.44} \\
    \textsc{Afraid}  & 0.14 & \textbf{0.16} & 0.15 \\
    \textsc{Sad}     & 0.35 & \textbf{0.40} & 0.39 \\
    \textsc{Amused}  & 0.20 & \textbf{0.22} & \textbf{0.22} \\
    \textsc{Happy}   & 0.29 & \textbf{0.32} & 0.31 \\
    \hline
    \textsc{Average} & 0.27 & \textbf{0.31} & 0.30 \\
    \hline
	\end{tabular} 		
	} 		 	
	\end{center}	 	
	\setlength{\belowcaptionskip}{-0.1cm} 	
	\caption{Pearson's correlation on \texttt{Corriere} left-out set.} 
	\label{tab:Corrieredev}
\end{table}

Furthermore, Tables~\ref{tab:Rapplerdev} and \ref{tab:Corrieredev} show the results obtained using all three versions of our resource, measured by Pearson correlation between the emotion annotation and the computed value (as indicated before).
When possible, obtained values are compared to those of %the previous approach of 
\texttt{DepecheMood$_{2014}$}; in general, it can be seen that the current work improves the performance with respect to the earlier version by a significant margin (6 points on average).

Considering the na\"{\i}ve approach we used, we can reasonably conclude that the quality and coverage of our resource are the reason of such results, and that adopting more complex approaches (i.e. compositionality) can possibly further improve the performance of %text-based 
emotion recognition.

\section{Evaluation}
\label{sec:experiments}

In order to thoroughly evaluate the generated lexica, we assess their performance in both regression and classification tasks.
The methodology described in Section~\ref{sec:depechemood++}, used to  obtain emotion scores for a given sentence/document, is common to all these experiments. The experiments are also restricted to the English \texttt{DM++}, as we are not aware of Italian datasets for emotion recognition. 

For all the experiments in this section we used the best \texttt{DM++} configuration found in the previous section, i.e. filtering out untagged documents and using a frequency cut-off set at 10. 

We report the performance obtained on the datasets described in Table~\ref{tab:datasets-reference}, and compare such results with the relevant previous works. Furthermore, in Table~\ref{tab:coverage}, we compare the coverage statistics over the same datasets for \texttt{DM$_{2014}$} and \texttt{DM++} (both with and without frequency cut-off). As can be seen, using \texttt{DM++} with cut-off 10 still grants a significantly higher coverage with respect to \texttt{DM$_{2014}$}, without losing too much coverage as compared to the version without cut-off.
% I CONSIDER IT DONE
%\textbf{HERE WE PUT A TABLE, HEADER SMT LIKE: NAME, DOMAIN, REFERENCE -- OR A BULLET LIST, THEN FOR EACH WE MAKE A SUBSECTION}

\begin{table*} [htb!] 	
	\setlength{\tabcolsep}{4pt} %shrink column size
	\begin{center} 	 	
		{\footnotesize 		
			\begin{tabular}{lcc} 		
				\hline 

	\textbf{Name} & \textbf{Domain}  &  \textbf{Reference}  \\
    \hline
    SemEval-2007 & News headline & \cite{strapparava2007semeval}\\
    Tweet Emotion Intensity  & \multirow{2}{*}{Twitter messages} & \multirow{2}{*}{\cite{mohammad2017wassa}} \\
    Dataset (WASSA-2017)  \\
    Blog & Blog posts & \cite{aman2007identifying} \\
    CLPsych-2016 & Mental health blogs & \cite{cohan2016triaging} \\
    \hline
	\end{tabular} 		
	} 		 	
	\end{center}	 	
	\setlength{\belowcaptionskip}{-0.1cm} 	
	\caption{Annotated public datasets used in the evaluation.} 
	\label{tab:datasets-reference}
\end{table*}

\subsection{Unsupervised Regression Experiments}
The SemEval 2007 dataset  on ``Affective Text"~\cite{strapparava2007semeval} was gathered for a competition focused on emotion recognition in over one thousand news headlines, both in regression and classification settings.
%\textcolor{violet}{REMOVE IF SPACE IS NEEDED:  Headlines typically consist of a few words and are often written with the intention to ``provoke" emotions, so to attract the readers' attention. An example of headline from the dataset is the following: ``\emph{Iraq car bombings kill 22 People, wound more than 60}". For the regression task the values provided are: \texttt{{\small $<$\textsc{Anger}(0.32),\textsc{Disgust}(0.27),\textsc{Fear}(0.84), \textsc{Joy}(0.0),\textsc{Sadness}(0.95),\textsc{Surprise}(0.20)$>$}} while for the classification task the labels provided are \{\textsc{Fear,Sadness}\}. Furthermore, this is to our knowledge the only dataset available providing numerical scores for emotions. }
This dataset was meant for unsupervised approaches (only a small development sample was provided) to avoid simple text categorization approaches.

% \begin{table} [h] 	
%	\begin{center} 	 	
%		{\footnotesize 		
%			\begin{tabular}{ll|ll} 						
%			\hline 
%            SemEval  & Rappler  & SemEval  & Rappler \\
%            			\hline 
%            \textsc{Fear} & \textsc{Afraid} & \textbf{\textsc{Suprise}} & \textbf{\textsc{Inspired}} \\
%            %ANGER & ANGRY & DISGUST& \textbf{ANNOYED}\\
%            \textsc{Anger} & \textsc{Angry} & - & \textsc{Annoyed}\\
%            \textsc{Joy} & \textsc{Happy} & - & \textsc{Amused}\\
%            \textsc{Sadness} & \textsc{Sad} & - & \textsc{Don't care}\\
%            \hline
%			\end{tabular} 		
%		} 		 	
%	\end{center}	 	
%	\setlength{\belowcaptionskip}{-0.1cm} 	
%	\caption{Mapping of Rappler labels on SemEval2007. In bold, cases of suboptimal mapping.}
%	\label{tab:mapping} 
%\end{table}  

\begin{table}[!htbp] 	
	\begin{center} 	 	
		{\footnotesize 		
		\setlength{\tabcolsep}{3pt} %shrink column size
			\begin{tabular}{c|c|ccc|ccc} 						
        %\texttt{lemma\#PoS} per headline & 5.41 \\
        %$M_{WE}$ \texttt{lemma\#PoS} per headline & 4.78 \\ 
        \hline
        %\textbf{Lexicon} & \textbf{Lexicon size} & \textbf{SemEval} & \textbf{WASSA17} & - & \textbf{CLPsych16} \\
        \multirow{3}{*}{\textbf{Dataset}} & \multirow{3}{*}{\textbf{DM$_{2014}$}} & \multicolumn{6}{c}{\textbf{\texttt{DepecheMood++}}} \\
        \cline{3-8}
        &&\multicolumn{3}{c}{\textbf{1 cutoff}} & \multicolumn{3}{|c}{\textbf{10 cutoff}} \\
        \cline{3-8}
        & & \textbf{tok} & \textbf{lem} & \textbf{l\#p} & \textbf{tok} & \textbf{lem} & \textbf{l\#p} \\
        \hline
        SemEval07 & 0.64 & 0.91 & 0.94 & 0.92 & 0.85 & 0.88 & 0.85 \\
        WASSA17 & 0.40 & 0.66 & 0.62 & 0.65 & 0.56 & 0.52 & 0.53 \\
        Blog posts & 0.64 & 0.93 & 0.92 & 0.91 & 0.84 & 0.83 & 0.80 \\
        CLPsych16 & 0.57 & 0.87 & 0.87 & 0.87 & 0.81 & 0.78 & 0.76 \\
        \hline
			\end{tabular} 		
		} 		 	
	\end{center}	 	
	\setlength{\belowcaptionskip}{-0.1cm} 	
	\caption{Statistics on words coverage per headline. tok: tokens, lem: lemma, l\#p: lemma and PoS.} 	
	\label{tab:coverage} 
\end{table} 

It is to be observed that the affective dimensions present in the test set -- based on the six basic emotions model~\cite{Ekman1971} -- do not exactly match with the ones provided by \texttt{Rappler}'s Mood Meter; therefore, we adopted the mapping previously proposed in~\cite{staiano2014depechemood} for consistency. 

In Table~\ref{tab:SemEval2007} we report the results obtained using our lexicon on the SemEval 2007 test data set. \texttt{DM++} is found to consistently improve upon the previous version (\texttt{DM$_{2014}$}) on all emotions, granting an improvement over the previous SOTA of up to 6 points.

\begin{table} [!htbp] 	
	\begin{center}{
	\footnotesize 		
	\begin{tabular}{l|c|ccc} 		
	\hline 
	\textbf{Emotion} & \textbf{DM$_{2014}$} &  \textbf{token}  &  \textbf{lemma} & \textbf{lemma\#PoS}  \\
    \hline
    \textsc{Anger}       & 0.37 & \textbf{0.47} & \textbf{0.47} & 0.44 \\
    \textsc{Fear}        & 0.51 &\textbf{0.60} & 0.59 & \textbf{0.60} \\
    \textsc{Joy}         & 0.34 & \textbf{0.38} & 0.37 & 0.35 \\
    \textsc{Sadness}     & 0.44 & 0.46 & 0.48 & \textbf{0.50} \\
    \textsc{Surprise}    & 0.19 & 0.21 & \textbf{0.23} & 0.22 \\
    \hline
    Average     & 0.37 & \textbf{0.43} & \textbf{0.43} & 0.42 \\
    \hline
	\end{tabular} 		
	} 		 	
	\end{center}	 	
	\setlength{\belowcaptionskip}{-0.1cm} 	
	\caption{Pearson's correlation on SemEval2007 dataset.}
	\label{tab:SemEval2007}  
\end{table}

% To further evaluate the performance we can obtain with our lexicon, we use .
%It is worth noting that, in average, our generated resource cover 72\% of the headline words of the dataset, which is superior to the previously reported coverage of 64\%~\cite{staiano2014depechemood} (see Table~\ref{tab:coverage}).

%Only one test headline contained exclusively words not present in \texttt{DepecheMood}, further indicating the high-coverage nature of our resource.
%In Table~\ref{tab:coverage} we report the coverage of some Sentiment and Emotion Lexica of different sizes on the same dataset. %(the average length of headlines is 7.37 words). 
%Similar to \cite{warriner2013norms}, we observe that even if the number of entries of our lexicon is far lower than that of resources such as SentiWordNet, the fact that we extracted and annotated words from documents grants a high coverage of language use.

\subsection{Supervised Regression Experiments}
\label{subsec:supervised}

Turning to evaluation on supervised regression tasks, the approach we use is inspired by~\cite{beck2014joint}:
% we devised a simple yet effective method that does not use \ac{bow} representations of sentences and it is partially drawn from the idea presented in \cite{beck2014joint}.
we cast the problem of predicting emotions %from text
as multi-task instead of single-task, e.g. rather than only using happiness scores to predict happiness, we use also the scores for sadness, fear, etc.

% In this work, the prediction task of an emotion is cast as a multi-task approach rather than a single-task one (e.g. instead of using  happiness scores alone to predict happiness, use also sadness, fear, etc.).
This approach is justified by the evidence that emotion scores often tend to be correlated or anti-correlated (e.g. joy and surprise are correlated, joy and sadness are anti-correlated).

Hence, for each dataset we built $N$ prediction models (one for each emotion present in the dataset, using the lexicon scores computed using either tokens, lemmas, or lemma\#PoS),
% with 
% the three pre-processing strategies (tokens, lemma, lemma\#PoS) 
as features:
% Therefore, depending on the number $N$ of emotions present in a dataset, we build $N$ models of the kind:
% So, for example in \texttt{corriere.it} we have built 5 models of the kind: 
\begin{equation}
e_i \sim \sum_{i=1}^{N} lex_i    
\end{equation}
where $e_i$ is the predicted score on emotion $i$, and $lex_i$ is the average score (computed on the elements of the test title/sentence) derived from the lexicon for emotion $i$.

% \texttt{\small{$<$ \textsc{Annoyed} $\sim$ \textsc{Annoyed}$_{pred}$ + \textsc{Afraid}$_{pred}$ + \textsc{Sad}$_{pred}$ + \textsc{Amused}$_{pred}$ + \textsc{Happy}$_{pred}$ $>$}}

We have used several learning algorithms: linear regression, \ac{svm}, decision tree, random forests and multilayer perceptron in a ten-fold cross-validation setting. 
Again, consider that we are not trying to optimize the aforementioned models, but just trying to understand if there is room for strong \textit{supervised} baselines that uses standard machine learning methodologies on top of our simple features (i.e. feeding the \texttt{DepecheMood++} lexicon scores to each emotion model). {Table~\ref{tab:supervised} shows that this is the case, with improvements ranging from 8 to 15 points, depending on the dataset.}

\begin{table} [htb!] 	
	\begin{center}{
	\setlength{\tabcolsep}{2pt} %shrink column size
	\footnotesize 		
	\begin{tabular}{l|l|ccccc|c} 		
    \hline
	\textbf{Dataset} & \textbf{Word rep.} & \textbf{LR} & 
	\textbf{SVM}  &  \textbf{DT} & \textbf{RF} & \textbf{MLP} & \textbf{DM++} \\
    \hline
    \multirow{3}{*}{Rappler}    & token         & 0.38 & 0.38 & 0.37 & \textbf{0.40} & 0.38 & 0.33 \\
                                & lemma         & 0.37 & 0.37 & 0.36 & \textbf{0.40} & 0.39 & 0.33 \\
                                & l\#p          & 0.36 & 0.36 & 0.35 & \textbf{0.38} & \textbf{0.38} & 0.33 \\
    \hline 		
    \multirow{3}{*}{Corriere}   & token         & 0.35 & 0.35 & 0.34 & \textbf{0.39} & 0.37 & 0.27 \\
                                & lemma         & 0.36 & 0.36 & 0.35 & \textbf{0.40} & 0.38 & 0.31 \\
                                & l\#p          & 0.35 & 0.36 & 0.35 & \textbf{0.40} & 0.37 & 0.30 \\
    \hline 		
    \multirow{3}{*}{SemEval}    & token         & 0.49 & 0.49 & 0.49 &\textbf{0.53} & 0.46 & 0.43\\
                                & lemma         & 0.50 & 0.50 & 0.48 & \textbf{0.53} & 0.47 & 0.43 \\
                                & l\#p          & 0.49 & 0.49 & 0.50 & \textbf{0.53} & 0.48 & 0.42 \\
    \hline 		
	\end{tabular} 		
	} 		 	
	\end{center}	 	
	\setlength{\belowcaptionskip}{-0.1cm} 	
	\caption{Regression results in supervised settings: Pearson's correlation averaged over all emotions. %l\#p: lemma\#PoS, 
	LR: linear regression, DT: decision trees, RF: random forest, MLP: multilayer perceptron.}
	\label{tab:supervised}  
\end{table}

Additionally, we have performed another experiment that compares \texttt{DepecheMood++} to existing approaches on a popular emotion dataset (Tweet Emotion Intensity Dataset): we replicated the approach outlined in~\cite{mohammad2017wassa} on the dataset therein presented, using \texttt{DepecheMood++} as lexicon.
% For this end, the same dataset as in has been used, and the prediction method has been replicated, using as lexicon \texttt{DepecheMood++}.

The results are reported in Table~\ref{tab:comparison1}:
% shows the results of the comparison of the use of \texttt{DepecheMood++} in the task of predicting emotion intensity in Twitter messages, as explained in~\cite{mohammad2017wassa}.
replicating the original work, Pearson's correlation has been used to measure the performance; since \texttt{DepecheMood++} is not a Twitter-specific lexicon, we also report non-domain-specific approaches.

% \textbf{NOT CORRESPONDING TO TABLE: It can be seen that \texttt{DepecheMood++} improves over previous in the emotions \textit{fear} and \textit{sad}, while the lexicon from~\citet{hu2004mining} performs better in \textit{anger} and \textit{joy}.} \textcolor{violet}{
It can be seen that \texttt{DepecheMood++} improves over \texttt{DM$_{2014}$} on all emotions, while the lexicon from~\citet{hu2004mining} performs slightly better only on \textit{joy}. In an aggregate view of the problem (average of all emotions) \texttt{DepecheMood++} yields the best performance among the non-Twitter-specific lexica.
% }

\begin{table} [htb!] 	
    \setlength{\tabcolsep}{5pt} %shrink column size
	\begin{center}{
	\footnotesize 		
	\begin{tabular}{l|cccc|c} 		
	\hline 
	\textbf{Lexicon} & \textbf{Anger} &  \textbf{Fear}  &  \textbf{Joy} & \textbf{Sad} & \textbf{Avg.}  \\
    \hline
    Bing Liu        & 0.33 & 0.31 & \textbf{0.37} & 0.23 & 0.31 \\
    MPQA            & 0.18 & 0.20 & 0.28 & 0.12 & 0.20 \\
    NRC-EmoLex      & 0.18 & 0.26 & 0.36 & 0.23 & 0.26 \\
    SentiWordNet    & 0.14 & 0.19 & 0.26 & 0.16 & 0.19 \\
    \hline
    DM$_{2014}$      & 0.35 & 0.28 & 0.27 & 0.30 & 0.30 \\
    \hline
    DM++ token     & 0.33 & \textbf{0.32} & 0.34 & 0.40 & 0.33 \\
    DM++ lemma      & \textbf{0.36} & \textbf{0.32} & 0.36 & \textbf{0.42} & \textbf{0.35} \\
    DM++ lemma\#PoS   & 0.31 & 0.31 & 0.34 & 0.41 & 0.34 \\
    \hline 		
	\end{tabular} 		
	} 		 	
	\end{center}	 	
	\setlength{\belowcaptionskip}{-0.1cm} 	
	\caption{Comparison with generic lexica of \cite{mohammad2017wassa} in the task of emotion intensity prediction. Avg. is the average over all four emotions.}
	\label{tab:comparison1}  
\end{table}

\subsection{Supervised Classification Experiments}
\label{subsec:classification}

Finally, akin to Section~\ref{subsec:supervised} but this time in a classification setting, we performed additional experiments to benchmark \texttt{DepecheMood++} against existing works in the literature.

In a first experiment, we replicated the work described in~\cite{aman2007identifying}, tackling emotion detection in blog data.
Consistently with the original work detailed in~\cite{aman2007identifying}, we trained Naive Bayes and Support Vector Machines models using only the \texttt{DepecheMood++} affective scores.
This serves as comparison to the original work, where \texttt{General Inquirer} and \texttt{WordNet-Affect} annotations were used. Results are reported in Table~\ref{tab:comparison2}, including the performance of the original \texttt{DM$_{2014}$} lexicon for comparison purposes.
We observe that \texttt{DepecheMood++} brings a significant improvement and outperforms the other models.

\begin{table}[htb!] 	
	\begin{center}{
	\footnotesize 		
	\begin{tabular}{l|cc} 		
	\hline 
	\textbf{System} & \textbf{Naive Bayes} &  \textbf{SVM}    \\
    \hline
    \cite{aman2007identifying}       & \multirow{2}{*}{72.08} & \multirow{2}{*}{73.89}  \\
    General Inquirer + WN-Affect \\
    \hline
    DM$_{2014}$       & 75.86 & 76.89 \\
    \hline
    DM++ token        & 76.01 & \textbf{78.04}  \\
    DM++ lemma          & \textbf{76.50} & 78.03  \\
    DM++ lemma\#PoS     & 74.18 & 76.18 \\
    \hline 		
	\end{tabular} 		
	} 		 	
	\end{center}	 	
	\setlength{\belowcaptionskip}{-0.1cm} 	
	\caption{Comparison with \cite{aman2007identifying} in the task of emotion classification accuracy.}
	\label{tab:comparison2}  
\end{table}

Further, we replicated the work detailed in~\cite{cohan2016triaging}, and assessed and compared with it the performance of \texttt{DepecheMood++} on a domain-specific task:
in~\cite{cohan2016triaging}, the authors tackle relevance prediction of blog posts on medical forums dedicated to mental health.
It is worth noting that as~\cite{cohan2016triaging} used the original \texttt{DepecheMood$_{2014}$}, this experiment also serves the purpose of evaluating the improvements brought by \texttt{DepecheMood++}, which are shown in Table~\ref{tab:comparison3}.
% The results of this experiment are contained in Table~\ref{tab:comparison3}, indicating again significant improvements .
% Again, our current approach improves over reported results, indicating that adding more data to the process of generating a lexicon effectively improves the generated annotations.

\begin{table} [htb!] 	
	\begin{center}{
	\footnotesize 		
	\begin{tabular}{l|cc} 		
	\hline 
	\textbf{System} & \textbf{Accuracy} &  \textbf{F1-macro}    \\
    \hline
    \cite{cohan2016triaging}  & \multirow{2}{*}{81.62} & \multirow{2}{*}{75.21}  \\
    \hspace{1em}using DM$_{2014}$ \\
    \hline
    DM++ token        & 82.15 & 81.34  \\
    DM++ lemma          & \textbf{82.26} & \textbf{81.46}  \\
    DM++ lemma\#PoS     & 82.25 & 81.44 \\
    \hline 		
	\end{tabular} 		
	} 		 	
	\end{center}	 	
	\setlength{\belowcaptionskip}{-0.1cm} 	
	\caption{Comparison with \cite{cohan2016triaging} in the task of mental health post classification.}
	\label{tab:comparison3}  
\end{table}

\section{Discussion of Results}
Several findings that are consistent across the datasets emerge from the above experiments. 
% OSCAR: I havent done statistical tests :(
%These are all statistically significant\footnote{To test significance we use the approach described in \textcolor{blue}{\textbf{Silver, Hittner, and May (2003) (which paper?)}}, a modification of Fisher's z transformation for comparing two overlapping correlations based on dependent groups~\cite{dunn1969correlation}.}.

%\subsection{Dataset size improve performances}

%\textit{\textbf{PREVIOUS TEXT:} For SemEval we have a general improvement in performances,  DM$_{new}$ improves (with statistically significant differences) over DM for 3 emotions out of 5 (namely ANGER, FEAR, SADNESS).
%For Rappler test-set we have a general improvement in performances as well.
%DM$_{new}$ improves (with statistically significant differences) over DM for 4 emotions out of 8 (namely AMUSED, ANGER, DONT\_CARE, INSPIRED).
%This comparison allow us to measure the effect of the dataset size (i.e. the new dump).}

As shown in Tables~\ref{tab:Rapplerdev} and~\ref{tab:SemEval2007}, the new version of  \texttt{DepecheMood++} effectively and consistently improves (see the additional benchmarks reported in  Tables~\ref{tab:comparison1},~\ref{tab:comparison2} and~\ref{tab:comparison3}) over the original work~\cite{staiano2014depechemood}.
% Similar comparisons have been also done in Tables~\ref{tab:comparison1},~\ref{tab:comparison2} and~\ref{tab:comparison3}.
Such improvements can be explained by the expansion of the training data, which enables the generated lexicon to better capture emotional information; in Figure~\ref{fig:tokenlemma}, we showed the performance obtained by lexica built on random and increasing subsets of the source data, and observe consistent improvements until a certain saturation point is met. 
% this 
%  effect is also visible in Figure~\ref{fig:tokenlemma}, where the lexicon generation approach is executed on dataset slices of varying sizes; also, to observe a saturation point after  which the improvement saturates, indicating that further expansions of the training data would not lead to more improvements.

%\subsection{Pre-processing has an effect on results}

%\oa{This is radically different to what it said before.}\textit{\textbf{PREVIOUS TEXT:} On the SemEval dataset we see that TOKENS performs as well as DM$_{new}$ and in one case even better (SADNESS).For Rappler test set, TOKENS in general perform as well as  DM$_{new}$ for 5 emotions out of 8, better in one case (SADNESS) and worse in two cases (ANGER,INSPIRED).}

Moreover, we found that adding a word frequency cutoff parameter leads to a benefit in the performance of the generated lexicon; in our experiments we find an optimal value of 10 for both the English and Italian lexica. 
% It has already been shown that the optimal value differs in the Rappler and Corriere dataset;
% this suggests that cutoff selection should be done dataset-wise.
% \citet{pang2002thumbs} discuss more on this, considering the cutoff strategy as a feature selection approach.

Turning to the benefits of common pre-processing stages, our experiments included tokenization, lemmatization and \ac{pos} tagging. While the original \texttt{DM$_{2014}$} lexicon only provided a lemma\#PoS-based vocabulary, we show that -- for English -- tokenization suffices, and further stages in the pre-processing pipeline do not significantly contribute to the generated lexicon precision; conversely, we obtained significant improvements by adding a lemmatization stage for Italian (see Figure~\ref{fig:tokenlemma}), a fact we hypothesize due to morphologically-richer nature of the Italian language.

% using either tokens or lemmas offer better results that lemmatization with \ac{pos} tagging, as shown in Figure~\ref{fig:tokenlemma}.
% Although many works tackle this issue, it is still an open question which pre-processing configuration is more beneficial, and normally this selection is done in an ad-hoc fashion.
% In our experimentation, we observe that this selection needs to take into account specific language characteristics, such as morphology (as seen in Sect.~\ref{subsec:unsupervised}). 

% As an additional comment, the computational cost of  the three pre-processing variations is not equal, being lemmatizing with \ac{pos} tagging the most expensive, and tokenization the cheapest one.
% Therefore, when using the generated resource, we recommend to use the approach that can yield good results and, at the same time, reduce the overall computational cost. For English, that would be using tokens, while in Italian it is more recommendable to use lemmas.
Further, as shown in Table~\ref{tab:emotional_documents}, filtering out untagged documents contributes to lexicon precision, arguably resulting in a higher-quality resource.
% Another interesting finding, drawn from Table \ref{tab:emotional_documents}, is that removing documents that show no emotion effectively improves the final performance of the lexicon.
% As discussed, such result strongly indicates that eliminating these non-emotional documents removes noise from the lexicon generation process, resulting in a higher-quality resource.

Finally, the extensive experiments reported in the previous sections show the quality of the English lexicon we release, in diverse domains/tasks. Our results indicate that additional data would not lead to further improvements, % can be obtained using additional data,
at least for English. Conversely, we note that the Italian resource we also provide to the community shows promising results.
% , and we emphasize the relevance of such release given the i) the scarcity of similar lexica and ii) the current inability to derive the Italian data since the source website has dismissed the emotional tags even from the archived articles.

\begin{figure*}[!htbp]
  \begin{subfigure}[b]{0.5\textwidth}
    \includegraphics[width=\textwidth]{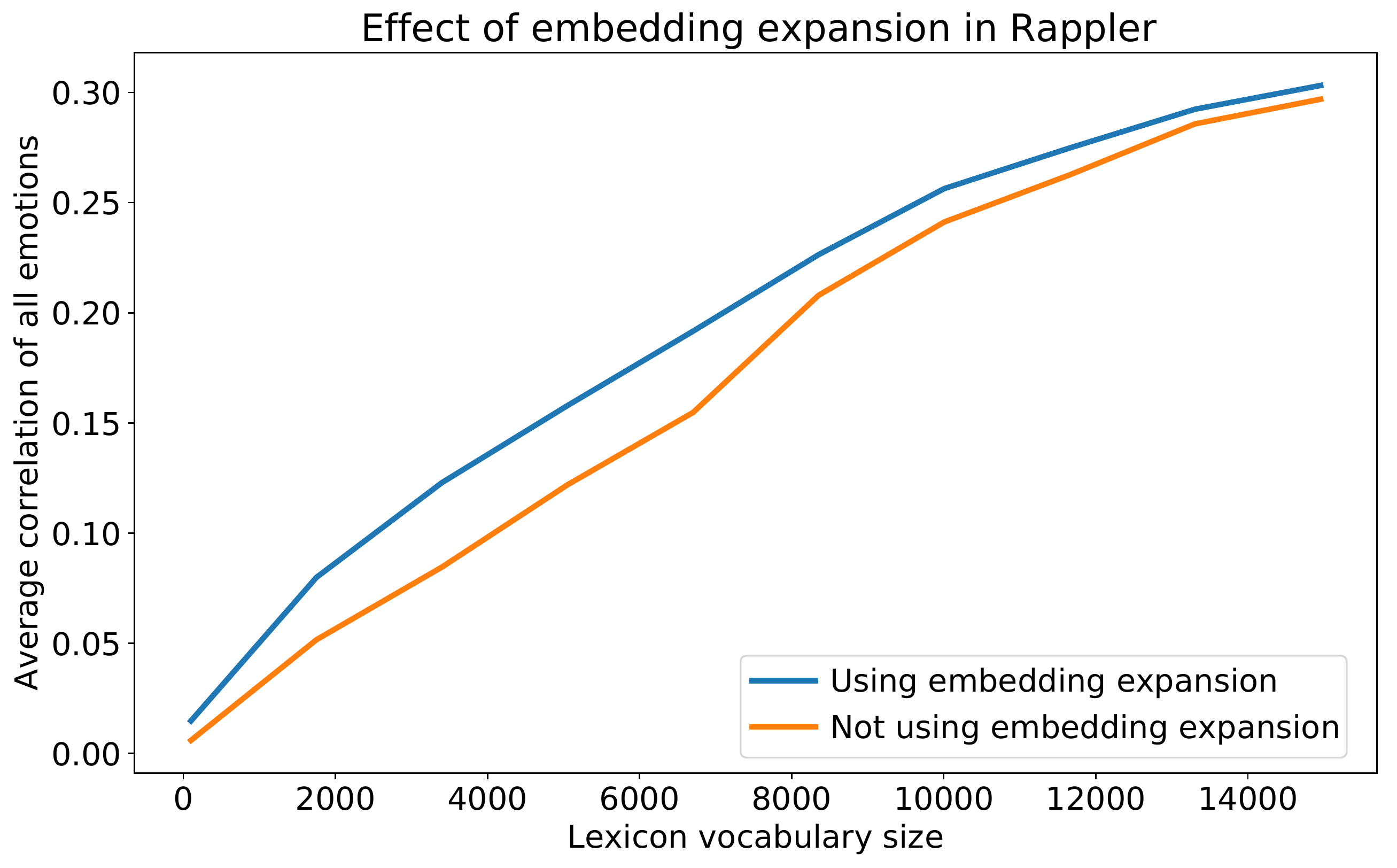}
    \caption{Rappler}
  \end{subfigure}
  \begin{subfigure}[b]{0.5\textwidth}
    \includegraphics[width=\textwidth]{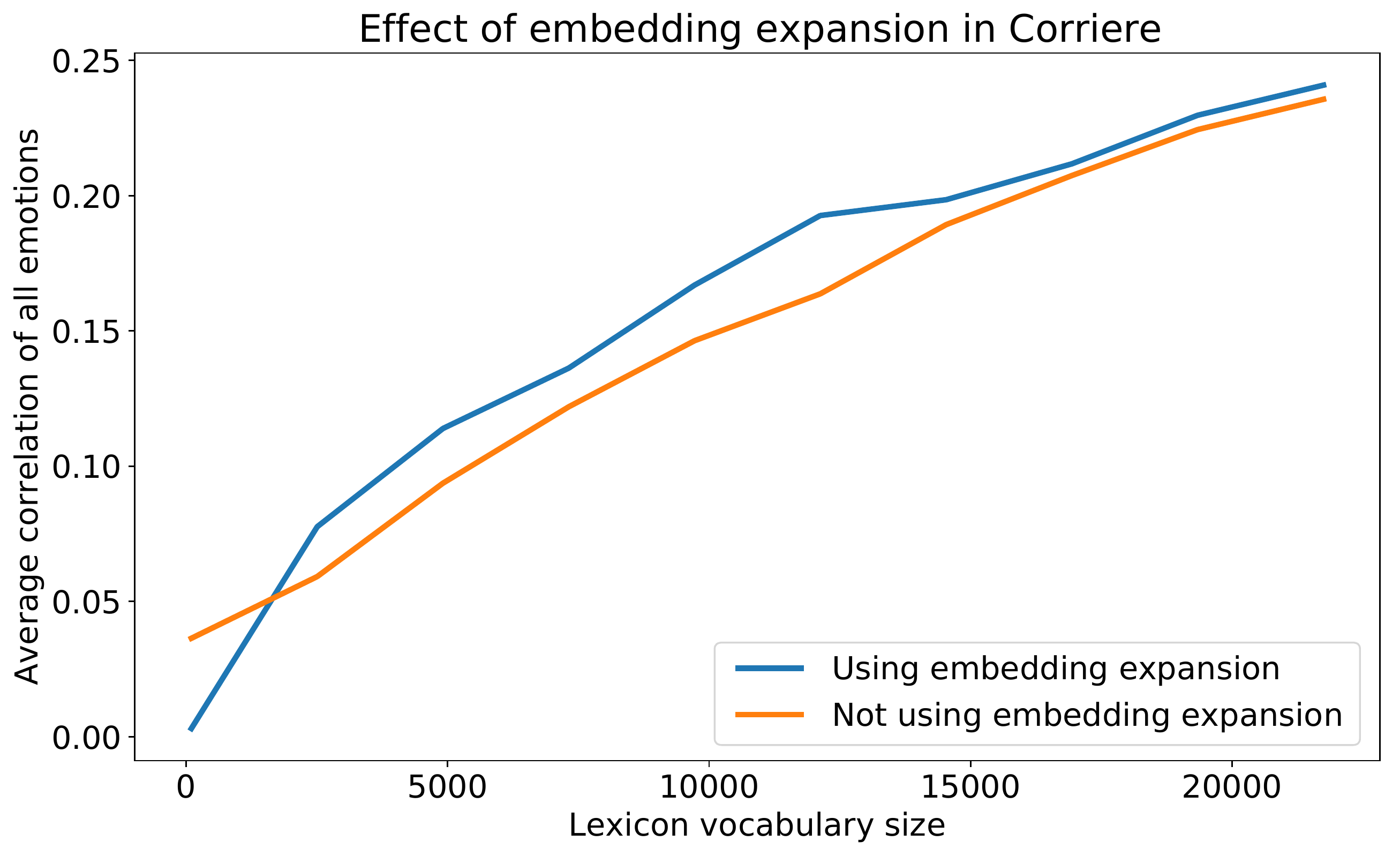}
    \caption{Corriere}
  \end{subfigure}
  
    \caption{Difference in performance when using the embedding expansion in both Rappler and Corriere.}
    \label{fig:emb-expansion}
\end{figure*}

\section{Increasing Coverage through Embeddings}
Over the last few years, word embeddings (dense and continuously valued vectorial representations of words) have gained wide acceptance and popularity in the research community, and have proven to be very effective in several NLP tasks, including sentiment analysis~\cite{GIATSOGLOU2017214}.

% {Finally, we tested a simple embeddings-based approach to further increase coverage of the generated lexica.
% In this final section we want to address an additional problem: while \texttt{DM++} leveraged a large data set of annotated news, we question if we can still increase lexicon coverage and performances Tstarting from smaller datasets.} In fact,
% a common limitation of sentiment and emotion lexica is that of vocabulary coverage: 
% }
%In fact, when using emotion lexica, a trade-off between relying on manual (high-quality) or automatically inferred (higher coverage) is to be found.

% In parallel to this, r

% Recent trends in Machine Learning applied to Natural Language Processing have appeared in the literature, outperforming previous systems~\cite{collobert2011natural} in many tasks, including sentiment and emotion analysis.
% This trend is mainly based on the use of a variety of neural networks architectures and of word embedding models, which basically offer continuous representations of words as vectors.
% Such representations have proven to be an effective technique in many NLP tasks, including sentiment analysis~\cite{GIATSOGLOU2017214}.

Taking into account this outlook, we propose a technique we call \textit{embedding expansion}, that aims to expand lexica vocabulary by means of a word embedding model.
The idea is to map words that do not originally appear in a certain lexicon to a word that is contained in the aforementioned lexicon.
%For this, the distance between words as measured by a given embedding model is used; 

Hence, given an out-of-vocabulary (OOV) word $w_i$, we search in the embedding space for the closest word that is included in the lexicon $l_j$ so that $d(w_i, l_j)$ is minimal, where $d(\cdot,\cdot)$ is the cosine distance in the embedding model.
Finally, the emotion scores from $l_j$ are assigned to $w_i$.

% In order to validate the proposed embedding trick, 
We have performed an evaluation over \texttt{Rappler} and \texttt{Corriere} datasets 
% with our proposed \texttt{DepecheMood++}.
using the token-based versions of \texttt{DepecheMood++}, 
with the aim of observing the effect of using embedding expansion. 

To this end, we proceed to:
\begin{enumerate}
    \item remove random subsets, of decreasing size, from the original lexicon vocabulary;
    \item apply the expansion at each step;
    \item measure performance against the corresponding test sets (i.e. the left-out sets described in Section~\ref{subsec:unsupervised}).
\end{enumerate}
% This is done over the totality of the original lexicon vocabulary: first removing a high percentage, and gradually reducing the aforesaid subset.
% In each step, the embedding expansion is applied.
The pre-trained word embeddings used for English are the ones published by \citet{mikolov2013efficient}, while for Italian we use those by \citet{tripodi2017analysis}; Figure~\ref{fig:emb-expansion} shows the results of this evaluation.

As observed, performing the embedding expansion can improve the performance of the emotion lexicon.
The higher improvement is achieved when the vocabulary has been reduced to roughly half of its elements for the two datasets.
When the vocabulary is not reduced, instead, the improvement tends to disappear.

Thus, we conclude that this improvement can enhance the performance of lexica with low coverage by expanding their vocabulary through an embedding model.
Nevertheless, when the lexicon has a high coverage (as in the case of  \texttt{DepecheMood++}), further extending it using the embedding expansion does not lead to meaningful improvements.

\section{Conclusions}
The contributions of this paper are two-fold: first, we release to the community two new high-performance and high-coverage lexica, targeting English and Italian languages; second, we extensively benchmark different setup decisions affecting the construction of the two resources, and further evaluate the performance obtained on several datasets/tasks exhibiting a wide diversity in terms of domain, languages, settings and task.

Our findings are summarized below.
\paragraph{Better baselines come cheap:} we have shown how straightforward classifiers/regressors built on top of the proposed lexica and without additional features obtain good performances even on domain-specific tasks, and can provide more challenging baselines when evaluating complex task-specific models; we hypothesize that such computationally cheap approaches might benefit any lexicon.

\paragraph{Target language matters:} we built our lexica for two languages, English and Italian, using consistent techniques and data, a fact that allowed us to experiment with different settings and cross-evaluate the results. In particular, we found that for English building a token-based vocabulary suffices and further pre-processing stages do not help, while for Italian our experiments highlighted significant improvements when adding a lemmatization step. We interpret this in light of the morphologically-richer nature of the Italian language with respect to English.

\paragraph{Embeddings do help (once again):} we investigated a simple embeddings-based extension approach, and showed how it benefits both performance and coverage of the lexica. We deem such technique particularly promising when dealing with very limited annotated datasets.
% In this paper we have investigated how simple and computationally cheap techniques can be used to improve both \textit{precision} and \textit{coverage} of a state-of-the-art lexicon  that has been automatically inferred from a dataset of emotionally tagged news.  To do so, we build upon the methodology proposed in~\cite{staiano2014depechemood}
% and released an extension of the original lexicon built on a larger dataset, as well as a novel emotion lexicon targeting the Italian language. To evaluate the quality of these lexica we performed experiments on several datasets/tasks exhibiting a wide diversity in terms of domain, languages, settings and task. The results we obtained show that:  training straightforward classifiers/regressors from our lexicon, allows us to obtain good performances also on domain-specific tasks, and provides much more challenging baselines for complex task-specific models. Furthermore, 
% depending on the morphological richness of the language, specific pre-processing steps (e.g. lemmatization) can be beneficial, but in general less pre-processing is required than the previous version of the resource. Finally we show that the coverage of an  under resourced lexicon can be extended using embeddings, and such technique proves to be beneficial in terms of performance.

\bibliography{Persuasive}
\bibliographystyle{acl_natbib_nourl}

\end{document}